%% file: main.tex
\documentclass[10pt, a4paper]{article}
\usepackage{template/lrec2022,amsmath,graphicx}
\usepackage{amssymb,amsfonts}
\usepackage{algorithmic}
\usepackage{textcomp}
\usepackage{hyperref}
\usepackage{makecell}
\usepackage{multibib}
\usepackage{titlesec}
\titleformat{\section}{\normalfont\large\bfseries\center}{\thesection.}{1em}{}
\titleformat{\subsection}{\normalfont\SmallTitleFont\bfseries\raggedright}{\thesubsection.}{1em}{}
\titleformat{\subsubsection}{\normalfont\normalsize\bfseries\raggedright}{\thesubsubsection.}{1em}{}
\renewcommand\thesection{\arabic{section}}
\renewcommand\thesubsection{\thesection.\arabic{subsection}}
\renewcommand\thesubsubsection{\thesubsection.\arabic{subsubsection}}
\usepackage{epstopdf}
\usepackage[utf8]{inputenc}
\usepackage{xstring}
\usepackage{soul}

\title{LibriS2S: A German-English Speech-to-Speech Translation Corpus}
\name{Pedro Jeuris and Jan Niehues} 

\address{Department of Data Science and Knowledge Engineering, Maastricht University \\
         Maastricht, The Netherlands\\
         p.jeuris@student.maastrichtuniversity.nl, jan.niehues@maastrichtuniversity.nl,\\}

\abstract{
Recently, we have seen an increasing interest in the area of speech-to-text translation. This has led to astonishing improvements in this area. In contrast, the activities in the area of speech-to-speech translation is still limited, although it is essential to overcome the language barrier. We believe that one of the limiting factors is the availability of appropriate training data. We address this issue by creating LibriS2S, to our knowledge the first publicly available speech-to-speech training corpus between German and English. \\
For this corpus, we used independently created audio for German and English leading to an unbiased pronunciation of the text in both languages. This allows the creation of a new text-to-speech and speech-to-speech translation model that directly learns to generate the speech signal based on the pronunciation of the source language. \\
Using this created corpus, we propose Text-to-Speech models based on the example of the recently proposed FastSpeech 2 model that integrates source language information. We do this by adapting the model to take information such as the pitch, energy or transcript from the source speech as additional input.
 \\ \newline }

\begin{document}

\maketitleabstract

\section{Introduction}
\label{sec:intro}
\input{01introduction}

\section{Related Work}
\label{sec:related}
\input{02relatedwork}

\section{Dataset creation}
\label{sec:data_aqui}
\input{05_dataAquisition}
\section{Source guided speech synthesis}
\label{sec:proposed}
\input{04_propsedmethod}
\section{Experimental setup}
\label{sec:setup}
\input{06_setup}
\section{Results}
\input{07_results}
\label{sec:results}
\section{Discussion}
\label{sec:disc}
\input{08_discussion}
\section{Conclusion}
\label{sec:conc}
\input{09_conclusion}
\section{Bibliographical References}\label{reference}
\input{references.tex}
\end{document}

%% file: 01introduction.tex
As the world becomes more and more connected, both online and offline, the need for language translation systems rises to overcome the large existing language barrier \cite{hutchins2003development,nakamura2009overcoming}. In the past few years the advances made within the Machine Translation (MT) domain have been significant. Computers have become better at translating text between two languages and can even translate speech to text in another language by using a single model \cite{bahdanau2016neural}. To solve the Speech-to-Speech Translation (S2ST) problem, in which a spoken phrase needs to be translated and spoken aloud in a target language, the problem is typically broken into three steps. In the first step the source speech gets transformed into text by using a Speech-to-Text (STT) model. The second step consist of translating the text to the target language. The last step takes in the translated text as input and synthesises speech in the target language using a Text-to-Speech (TTS) model. Because of the concatenation of these three steps, and the intermediate text representation, there is a loss of information from the original speech which does not get passed on after the first step \cite{sperber2020speech}. This information loss mainly consists of spoken language characteristics such as the pitch and energy in the source speech which are key elements for prosody. Prosody is essential for a conversation as it contains, for example, information on the emotional state of the speaker. 

\begin{figure}[t]
    \centering
    \includegraphics[width=\linewidth,height=2.6cm]{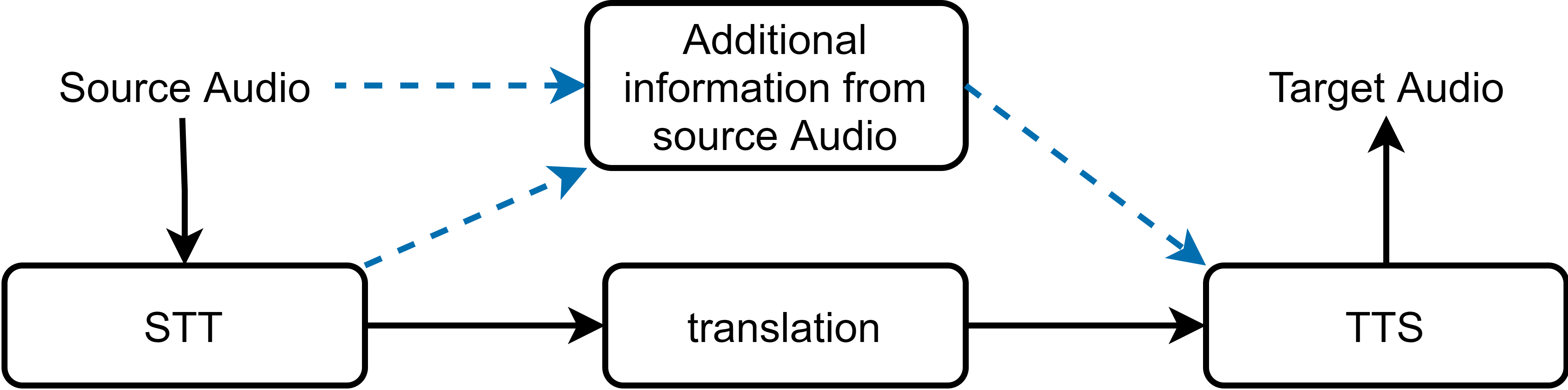}
    \caption{Conventional (black, full lines) and proposed (blue, dashed lines) additions to the pipelines.}
    \label{fig:pipeline}
\end{figure}

In this work, we want to address this issue by proposing dedicated TTS systems for speech translation that also take the source audio into account. 
The idea is that when synthesising the speech of a target language, information about the pitch and energy from the source language can be valuable to generate the right prosody. Of course, there is no one-to-one mapping. Therefore, we propose to integrate this information into the neural TTS system such that the model can automatically learn when to make use of the source language information. This is done by adding speech characteristics such as the pitch and energy from the speech in the source language into the TTS system as an additional input to the system as shown in Figure \ref{fig:pipeline}.


The main challenge we need to address is the creation of appropriate training data. In order to train the proposed model, we need to have aligned speech in both the source and target language. This parallel speech corpus should be created independently from each other to have high quality and unbiased audio in both languages. Therefore, we build the corpus based on the Librevox\footnote{https://librivox.org/} project. In this case, we aligned the audio of audiobooks in German and English. The speech in both languages was created independently of each other.


The contributions of this research are first of all the creation and release of LibriS2S\footnote{https://github.com/PedroDKE/LibriS2S} (Libri Speech-to-Speech), a speech aligned corpus between German and English with audio in both languages.

In addition, we made the scripts publicly available to encourage the creation of additional S2ST data sets.
With the help of these alignments and scripts a German-English speech aligned dataset can be made. By adapting the scripts this could be extended to more languages too. Secondly, we perform an investigation on how source text and audio features can be used to improve the quality of the synthesised speech from the TTS model. We propose an extension of the state-of-the-art TTS system FastSpeech 2 \cite{ren2021fastspeech} to integrate these features as well as evaluate the influence of the presented data.

%% file: 02relatedwork.tex
\subsection{Datasets}
Most datasets used in current state-of-the-art translation research such as LibrivoxDeEn \cite{beilharz19}, CoVoST2 \cite{wang2020covost}, MuST-C \cite{CATTONI2021101155} or LibriSpeech ASR\cite{librispeech} augmented with French translations \cite{kocabiyikoglu2018augmenting} are focused on Speech-to-Text translation. These datasets only contain the audio and transcription in the source language paired with the translation, in text, from the target language. Furthermore, datasets that do contain the audio from the target language such as Fisher and CALLHOME Spanish-English dataset \cite{post2013improved}, SP2 speech corpus \cite{inproceedings}, the EMIME Project \cite{kurimo-etal-2010-personalising}, VoxPopuli\cite{wang2021voxpopuli} and MaSS \cite{zanon-boito-etal-2020-mass}  either require a costly license, only have a limited amount of samples, exist out of unlabeled data or might have copyright concerns.\\
Since audio pairs of the same sentences in two languages are needed, to build the proposed TTS system and to provide the additional information during training and testing, this work will build further upon already existing speech-translation datasets. The missing speech segments from the target language will be scraped from the internet and aligned to match the dataset.

\subsection{S2ST and TTS systems}

\begin{figure*}[t]
    \centering
    \includegraphics[width=0.75\textwidth,height=8.6cm]{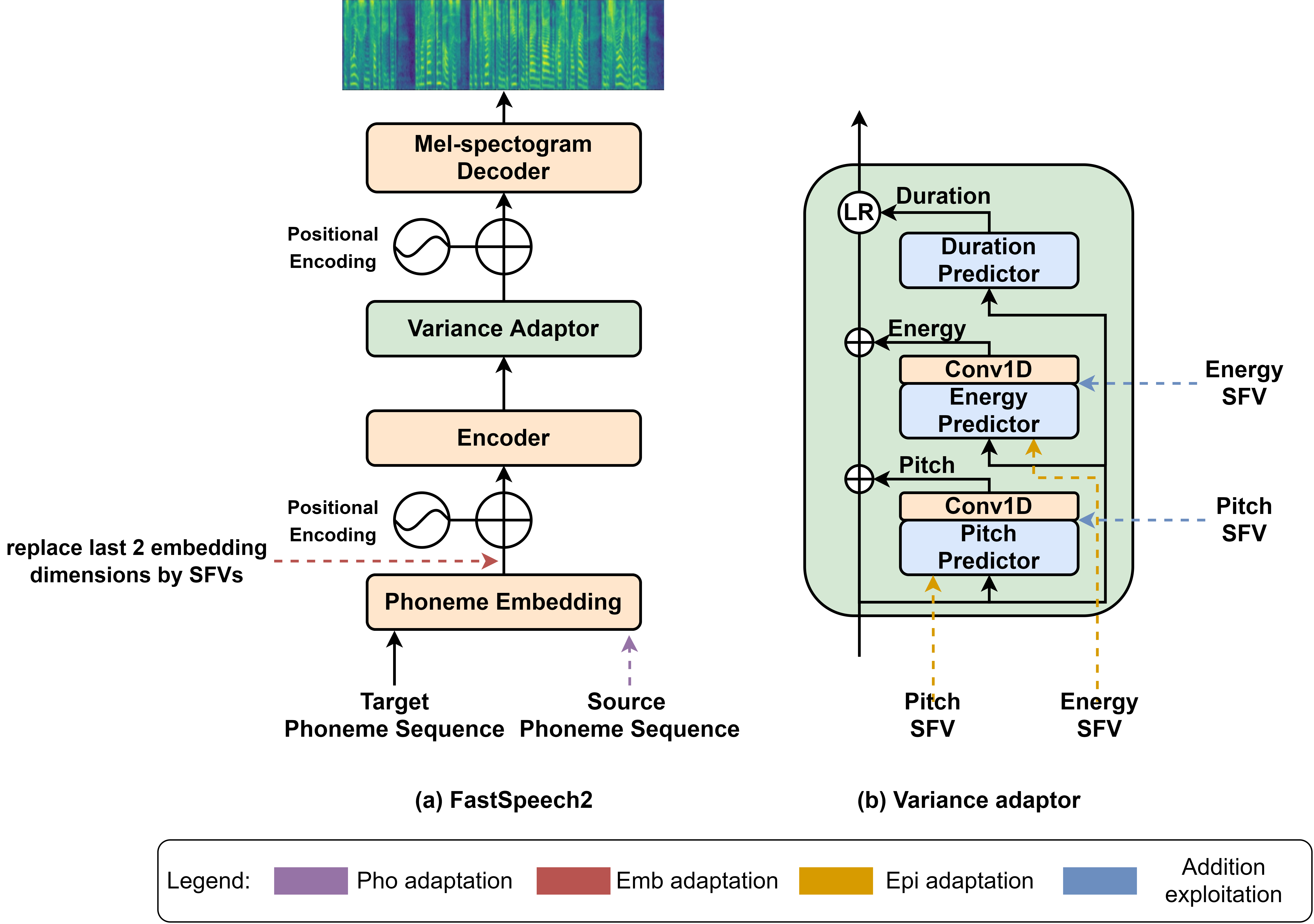}
    \caption{FastSpeech 2 architecture, figure adapted from \protect\cite{ren2021fastspeech}. Abbreviations: LR: Length Regulator (introduced in the original FastSpeech paper \protect\cite{ren2019fastspeech}, further information on the embedding, encoder, decoder and predictors can be found in the FastSpeech papers)}
    \label{fig:fs2arch}
\end{figure*}

Modern day S2ST systems such as provided by Jibbigo \cite{Jibbigo}, Microsoft Azure \cite{microsoftazure} and Skype \cite{lewis2015skype} make use of a concatenation of state-of-the-art models. As mentioned in the introduction this can result in a loss of information.
The oldest pipeline to take additional speech information, such as prosody, into account was that of Verbmobile \cite{wahlster2013verbmobil}. Within the Verbmobile project, pitch, energy and other features of speech were used to annotate the generated text with additional information and to use this additional information to help with the translation. Trying to achieve the same goal, in this paper the source information will be used only to enhance the quality of the generated speech at the end of the pipeline.

\newcite{porosody_estimate} try to incorporate emphasis in their TTS model with the help of an emphasis estimator. Their emphasis estimator takes the source text as input and predicts for each word an emphasis probability between 0 and 1. They then transfer this emphasis estimation to the target language and use these values as additional input of their TTS model. Compared to this work, source information will also be mapped to the target language but instead of using an intermediate predictor, the values are extracted from the source speech to be used. Also, instead of using a Hidden Semi-Markov (HSM) TTS Model, our approach uses a deep learning method.

The work of \newcite{skerryryan2018endtoend} focuses on transferring prosody between two speakers within the same language. By adapting the Tacotron \cite{wang2017tacotron} architecture to contain an extra prosody embedding layer they are able to transfer prosodic characteristics between speakers in the same model.

The feature generation model used in this work is the recently introduced FastSpeech 2 \cite{ren2021fastspeech}. This model is of particular interest because it has separate predictors for pitch and energy as can be seen in Fig. \ref{fig:fs2arch}. Compared to the original paper, the implementation\footnote{The baseline implementation used in this work can be found in the following GitHub repository: https://github.com/TensorSpeech/TensorFlowTTS} used has a few differences: 1) The pitch and energy are predicted per phoneme. 2) The pitch and energy are normalized using the z-normalization and these are being predicted. 3) Additionally a post-net similar to the one introduced in the Tacotron 2 \cite{shen2018natural} paper is used. This post-net is used to "predict a residual to add to the prediction to improve the overall reconstruction" \cite{shen2018natural}.\\

%% file: 05_dataAquisition.tex
\label{subsec:descrip_data}

In order to generate the most natural speech in the target language, we were targeting a data source where the source and target language speech was generated as naturally as possible. We found the Librivox project as a promising resource, as here the audio books in both languages are generated independently of each other. 

The dataset used as a starting block in this research is the LibrivoxDeEn \cite{beilharz19} dataset. This dataset consists out of German speech, transcriptions and their English translation of audiobooks. These audio books are all available on Librivox and their texts are also freely available. The English translations within this dataset are based on the official translations, which means that the English counterparts of the audiobook is possibly available on Librivox too.
Using the audiobooks from Librivox has multiple advantages:\\ 1) it is possible to collect parallel data with limited involvement of human aid, 2) some books have multiple speakers which could be used to construct for multi-speaker TTS systems, 3) to build further upon the previous advantage, this method also provides multiple-to-one speaker and one-to-multiple speakers alignments which could be interesting for future research, 4) there is only a limited number of legal and privacy concerns as all data used is part of the public domain according to Librivox and 5) a wide variety of languages are available to further facilitate multi-lingual S2ST.
\subsection{Data collection and allignment}
The first step to enhance the LibrivoxDeEn dataset was to download the English counterparts of the audiobooks from Librivox. As some of the audiobooks do not have the same amount of chapters across languages, for example the German audio book used in this research contains 29 chapters while the English audio book has 27, these were manually split or stitched together to use the texts given within the LibrivoxDeEn dataset.\\
The next step was to align the newly gathered English speech to the texts. This was done using the tool aeneas \cite{pettarin_2017}. Given the text and audio of a book's chapter that we want to align, this tool gives us the timestamps of when a phrase was said and makes it possible to extract the spoken sentences necessary to create the required data for S2ST research.
\subsection{Dataset analysis}
Using this method it was possible to align 12 audio books. From those 12 books, 8 did not require the additional prepossessing to align the chapters. Those alignments are made publicly available to download together with the tools used to scrape the audio from librivox and align the audio to their text.
\begin{table}[t]
\centering
\begin{tabular}{|l|l|l|}
\hline
& \textbf{German}   & \textbf{English}  \\ \hline
\# Audio files & 25 635   & 25 635   \\ \hline
\# unique Tokens & 10 367 & 9 322 \\ \hline
\# Words & 49 129 & 62 961 \\ \hline
\# Speakers & 42 & 29 \\ \hline
\begin{tabular}[c]{@{}l@{}}Duration\\ (hh:mm:ss)\end{tabular} & 52:30:57 & 57:20:10
\\ \hline
\end{tabular}
\caption{Collected data sizes}
\label{table:data_collected}
\end{table}\\
Additional metrics on the 12 books that were aligned can be found in Tab. \ref{table:data_collected}. To count the number of speakers these were counted separately for each book and some speakers can thus have been counted more than once.
The speakers that record the audio books on librivox do not always have access to a high-quality microphone. Because of this it can be challenging to create a TTS system for a single speaker. Moreover, the average length of audio per book, about 4.77 hours for English and 4.38 hours for German, is on the lower side when compared to the much used datasets LJSpeech \cite{ljspeech17} for English or Thorsten \cite{thorsten} for German with both over 23 hours of speech. However, using multiple speakers or audiobooks from the dataset to train a multi-speaker TTS model can be beneficial for the final audio quality \cite{chen2020multispeech} and makes it possible to make use of the full dataset. Furthermore, the scripts released with this work make it possible to align more audiobooks to the LibrivoxDeEn dataset. Combining both methods from this work and from \newcite{beilharz19}, used to create LibrivoxDeEn, it is possible to create speech aligned datasets for multiple languages given that the right audiobooks are available in both languages on librivox.

Using audio books does not give us a guarantee that sentences in both languages are expressed with the same prosody or emotion. Given the scarcity of such data, audio books can be used as a starting point as these have multiple advantages as discussed in the beginning of Sec \ref{subsec:descrip_data}.

%% file: 04_propsedmethod.tex
Given the newly created data we are now able to build TTS models that can use the audio from the source language to improve the generated speech in the target language. In this work we suggest models that use this information to help predict the energy and pitch of the target language. We do this by making three adaptations of the model that are then trained with the newly gathered data and one exploitation model where we try to improve the baseline without retraining. 
The different implementations are shown in Fig. \ref{fig:fs2arch} and further explained in this section.

\subsection{Source phoneme input}
The first adaptation takes a concatenation of both the source and target transcripts (in phonemes) as input. The embedding layer then maps both phoneme sequences to a vector space and then the encoder is used to mix the information from both embedded sequences.\\
This is the most straightforward integration, where only very limited adaptation to the existing architecture is needed. However, only source text information is passed to the model and the information is not encoded specifically making the task of using this additional information very difficult. Furthermore this representation does not give any additional information on the expressed prosody in the source speech.
This model will later on be referred to as 'pho' because of the additional phoneme input.

\label{subsec:add}
\begin{figure}[t]
    \centering
    \includegraphics[width=\linewidth, height = 5cm]{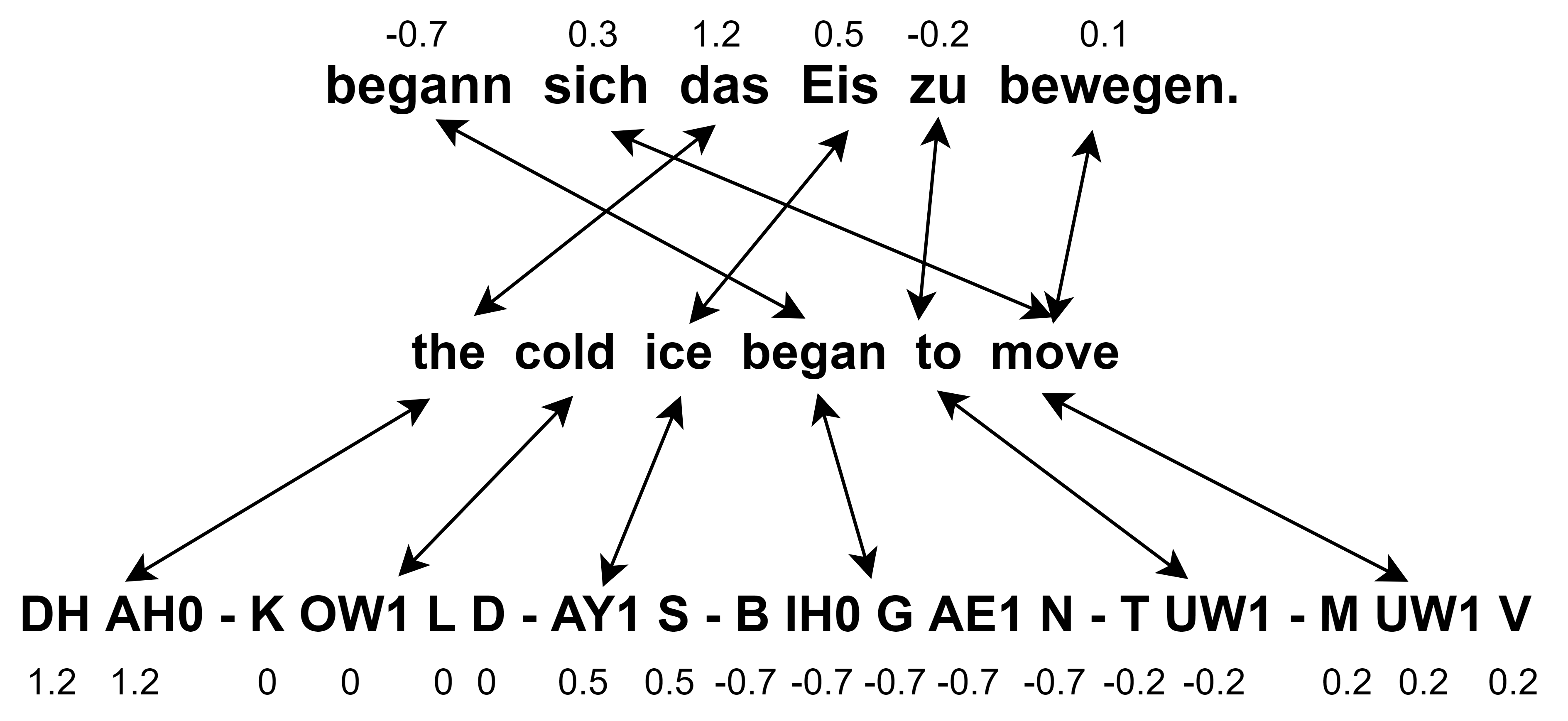}
    \caption{A (synthetic) example of an SFV to show the idea. Note how for the last word the average is being taken since it has a mapping from multiple words in the source}
    \label{fig:sfv}
\end{figure}

\subsection{Word level pitch and energy embedding}
\label{subseb:emb}
For this adaptation a Source Feature Vector (SFV) is being calculated and used to give the model additional information on the pitch and energy from the source speech.\\
The first step in creating the SFVs is to calculate the average pitch and energy for each word in the source speech. Afterwards with the help of multilingual alignment, where each word of a source sentence is mapped to their corresponding word in the translation, these SFVs can be mapped to the target language. This way a vector is created where for each phoneme in the target language either contains information on the average word-level value in the source language or, if it was not alligned, a zero value. An example of an SFV can be seen in Fig. \ref{fig:sfv}.\\
The second adaptation uses this additional information by replacing the last 2 dimensions in the embedding layer by the pitch and energy SFVs. This way the encoder takes the additional information from the SFVs as input which in turn can help the variance adaptor to predict the pitch and energy.\\
In the results section this model will be refered to as 'emb' because of the additional information that is incorporated in the dimensions of the embedding.

\subsection{Word level pitch and energy as additional input to predictor}
Instead of only using the SFVs as additional information to the embedding, the third adaptation builds further upon the previous introduced model in Sec. \ref{subseb:emb} and also uses the SFVs as additional input to their respective predictors. This way, the predictors also receive the values from their respective SFVs directly. This is done by stacking them to the encoder output to create a vector with sizes (2, \#PHONEME). This model is refered to as 'epi' which stands for Embedding and Predictor Input.
\begin{figure}[t]
    \centering
    \includegraphics[width=\linewidth,height=5cm]{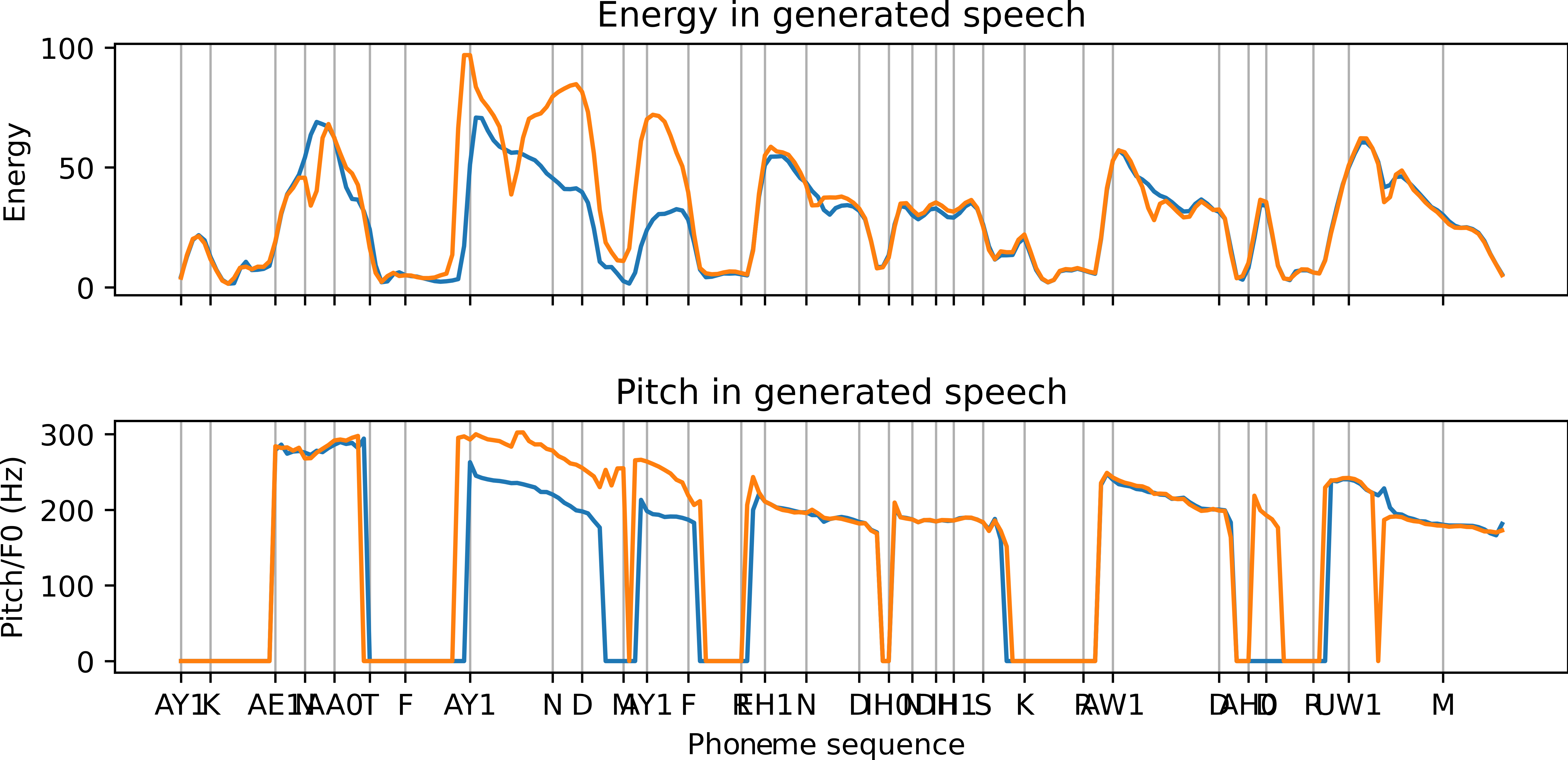}
    \caption{input sentence: "I cannot find my friend in this crowded room", "AY1 KAE1NAA0T FAY1ND MAY1 FREH1ND IH0N DHIH1S KRAW1DAH0D RUW1M"
    blue: using the output of the predictor, orange: increasing predicted energy and pitch at the phonemes corresponding to "find my"}
    \label{fig:influence_predictors_bad}
\end{figure}
\subsection{Add pitch and energy to predictor output}
Since the pitch and energy predictors try to estimate these values on a phoneme level, this can be exploited to influence the generated speech. As can be seen in Fig. \ref{fig:influence_predictors_bad} the energy and pitch in the generated speech can be changed by increasing the output values of their predictors before passing the values through the embedding. But it can also be seen that changing values at certain parts of the predictions can result in changes of pitch or energy in other parts of the generated speech too.\\ 
In the fourth model, this is being exploited within the baseline model by summing the SFV to the output of their respective predictor. This way phonemes from words that have a below average pitch or energy in the source language get their predicted values lowered (because of having a negative z-score added to them) and the same way words with an above average value in the source language get their values increased in the prediction.

\begin{table*}[t]
\centering
\begin{tabular}{|l|l|l|l|l|l|l|}
\hline
         &MOSNet prediction& {Pitch $\sigma$} & {Pitch $\gamma$} & {Pitch $\kappa$} & {Pitch DTW} & {Energy MAE}                   \\ \hline
GT       &3.673& 31.867                & 0.788                & 1.769                 & \textbackslash{} & \textbackslash{} \\ \hline
baseline &3.133& 41.163                & -1.138               & 2.627                 & 21.423 & 10.039          \\ \hline
pho      &3.161& 40.779                & -1.063               & 2.931                 & \textbf{19.876} & 10.110           \\ \hline
emb      &\textbf{3.163}& \textbf{38.113}       & -1.000               & 3.104                 & 20.329 & \textbf{10.002}            \\ \hline
epi      &3.159& 38.704                & -1.039               & 2.979                 & 19.948 &  10.103 \\ \hline
addition &3.071& 42.174                & \textbf{-0.807}      & \textbf{2.390}        & 23.065 &  11.042        \\ \hline
\end{tabular}
\caption{Predicted MOS scores, pitch moments/DTW and energy MAE from the GT and generated audio}
\label{table:evaluation_metrics}
\end{table*}

%% file: 06_setup.tex
\subsection{Preprocessing}
\label{subsec:prepro}
To convert the transcript to a phoneme sequence and extract the duration of each phoneme the Montreal Forced Alignment tool \cite{mfa} was used. The English sequences uses the ARPAbet phoneme set and the German sequences uses the prosodylab format. To get the cross-lingual alignment to create the SFVs the awesome-align tool\footnote{https://github.com/neulab/awesome-align} \cite{awesomealign} was used.\\
To train the TTS models, the speech data was resampled to 22050 kHz and the MFCC's were extracted using a hop-size of 256, a frame size of 1024 and with a basis of 80. For training, testing and validation only audio files with a length between 1s and 20s were used.
\subsection{Data size}
Some of the collected books either have bad microphone, need further processing to separate the speakers or do not have enough samples to train a single speaker system. Therefore in the continuation of this research the book Frankenstein by Mary Shelly\footnote{https://librivox.org/frankenstein-or-the-modern-prometheus-1818-by-mary-wollstonecraft-shelley/ and https://librivox.org/frankenstein-oder-der-moderne-prometheus-by-mary-wollstonecraft-shelley/} is used. As a result there were a total of only 2335 audio files used for this research. 2079 of these were used for training, 129 for validation and 127 for testing. This number of samples for training is on the low side compared to datasets usually used for single speaker TTS models such as LJSpeech \cite{ljspeech17} where a total of 13,100 English audio fragments are used. Nevertheless, it is still sufficient to train a TTS model but might result in more/faster overfitting and worse audio quality in the generated speech.
The book used also had some misalignments on the source side originating from the LibrivoxDeEn dataset but was realigned using the same method described in Sec. \ref{sec:data_aqui}.
\subsection{Evaluation}
\label{subsec:eval}
The vocoder model used in this research is the MB-MelGAN model \cite{mbmel} which has been fine tuned on the samples in the training set and then used to generate the 127 samples from the test set to evaluate the speech generated.
As getting a Mean Opinion Score (MOS) is not possible due to the inaccessibility to enough native speakers, the MOS is approximated by using the deep learning approach MOSNet \cite{mosnet}. The disadvantage about this method is that it is just an approximation and might not represent the real MOS that a human evaluation would give. Fortunately, this method also has advantages such as: more samples can be evaluated, no new biases are introduced during evaluation and the same model is used for testing which results in comparable metrics.
To evaluate the energy and pitch in the generated speech, the same methods as in the FastSpeech 2 \cite{ren2021fastspeech} papers are used. To evaluate the pitch, the standard deviation, skewness and kurtosis for the pitch moments in the generated speech is calculated \cite{pitch,pitch2}. Ideally this would be as similar as possible to our ground truth. Additionally the average Dynamic Time Warping (DTW) distance of the pitch is also computed with the help of the dtw-python package \cite{DTW}. To evaluate the energy, the MAE between the generated speech and the ground truth is being calculated. Similar to \newcite{ren2021fastspeech}, the durations extracted by the MFA are used in the length regulator to ensure the same duration in the generated audio as the ground truth to calculate the MAE of the energy in the generated speech.

%% file: 07_results.tex
\subsection{Speech metrics of generated samples}
The metrics achieved by evaluating the generated speech of all models can be seen in Tab. \ref{table:evaluation_metrics}. The first metric is the predicted MOS of the generated speech. According to the predicted score the three model adaptations improve over the baseline and result in a slightly more natural audio quality. The only model not improving on this metric is the addition exploitation of the baseline model.\\
The pitch moments from the generated speech can be found between the third and fifth column. Ideally these are as close as possible to the ground truth from the same test set. As can be seen, all trained model adaptations have an improvement on the baseline when looking at the standard deviation ($\sigma$) and skewness ($\gamma$) but perform worse on the resulting kurtosis ($\kappa$). The addition exploitation however achieves the best results on the skewness and kurtosis of the pitch but performs worse when looking at the standard deviation. Besides the pitch moments, the DTW distance to the GT is also given in the sixth column. We can see that again for the three adaptations this distance has decreased when compared to the baseline model indicating a better modeling of the pitch. For the addition model this distance has increased, indicating a worse modeling compared to the baseline. This is probably due to the effect shown in Fig. \ref{fig:influence_predictors_bad} where unwanted changes in the pitch can happen or because of the fact that prosody is not a one-to-one mapping across two languages.\\
For the MAE between the energy within the GT and generated speech we can see that only the emb adaptation improves over the baseline and the other two adaptations perform similar to each other. The addition model however performs the worst again, possibly due to the same reasons as to why the DTW distance increased for this model.

\subsection{Ablation study on the influence of SFV}
To have a closer look at the influence of the SFV on the pitch and energy in the generated speech an ablation study is done where the SFVs are set to zero before passing them to the emb and epi model during inference. Afterwards the same metrics as the previous three sections are calculated.
The values achieved with the SFVs extracted from the source speech are also given as GT in the table, these values are the same as from Tab. \ref{table:evaluation_metrics}.
\begin{table}[h]
\centering
\resizebox{\linewidth}{!}{
\begin{tabular}{|l|l|l|l|l|l|l|}
\hline
             &MOS & $\sigma$ & $\gamma$ & $\kappa$       & DTW          & MAE \\ \hline
\makecell[l]{emb\\zero sfv} & \textbf{3.193} & 40.187   & -1.039   & 3.114          & \textbf{20.094} & \textbf{9.985}     \\ \hline
\makecell[l]{emb\\GT sfv} &3.163 & 38.113   & -1.000   & 3.104          & 20.329 & 10.002    \\ \hline
\makecell[l]{epi\\zero sfv} & 3.153& 38.988   & -1.074   & \textbf{2.730} & 20.123          & 10.137       \\ \hline
\makecell[l]{epi\\GT sfv}  & 3.159 & 38.704   & -1.039   & 2.979 & 19.948          & 10.103           \\ \hline
\end{tabular}}
\caption{pitch and energy metrics from the generated speech with SFV values being zero. Improvements over non-zero SFV are shown in bold.}
\label{tab:zero_Sfv}
\end{table}
\\Tab. \ref{tab:zero_Sfv} shows the achieved metrics when the SFVs values are put to zero during inference. Due to the z-normalization a zero value means that this part has an average pitch or energy. For the epi model, setting the SFVs values to zero only results in the improvement on one metric. This indicates that the information from the SFV is used in the epi model and improves the final performance of the model. For the emb model it is not as clear and we only see improvements in half the metrics.
One reason could be that the SFVs are more directly integrated in the epi approach and therefore it is easier for the model to use this information to generate the MFCC.

%% file: 08_discussion.tex
Given the results presented in Tab. \ref{table:evaluation_metrics} and Tab. \ref{tab:zero_Sfv}, it can be said that while changing the output of the predictors might be a promising way to control the pitch and energy (as seen in Fig. \ref{fig:influence_predictors_bad}) in the generated speech, combining them with the SFVs fails to improve over the baseline on most metrics and has the overall worst performance out of the methods presented.
The emb model, where the SFVs are used as additional input to only the embedding space, does improve over the baseline and performs best out of the proposed methods on half of the metrics, but this method gives conflicting results in the ablation study and is therefore a less interesting method. Using the SFVs also as input to the predictors gives the second best results on all metrics and the ablation study shows that using this method can be beneficial for the epi model. 
While the pho model does not receive any additional information on the energy and pitch from the source speech, only information of what was said, this still seems to improve the generated speech. On most metrics it improves over the baseline and seems to perform similarly to the other two adaptations, emb and epi models, in terms of predicted MOS, DTW distance and the energy MAE.

%% file: 09_conclusion.tex
In this paper, LibriS2S, a dataset consisting of sentence aligned German and English audio pairs and their transcriptions, is introduced and released together with the methods used to make this dataset. The presented method has multiple advantages and is shown to be able to be used for a single-speaker TTS system. This dataset is also used to incorporate additional source information during training and inference of a FastSpeech 2 model in a S2ST environment. The trained models seem to mainly improve on having a better pitch characteristic as the target and have a higher predicted MOS score by using additional information from the source audio. On the energy level of the generated speech the adaptations do not seem to benefit much from the additional source information. For some of the adaptations conflicting results are achieved in the ablation study and thus might need further investigation. However, this is not the case for all models presented in this work.\\
The presented data opens up several interesting research directions. While the baseline experiments focused on a single speaker scenario, the provided data offers data from multiple speakers, which can be beneficial for the quality of generated speech\cite{multispeakertts}. The use of the full data will also enable the creation of a full speech translation pipeline and allow to investigate the challenges in modeling the full pipeline. 
Furthermore, the provided toolkit enables the creation of similar data for other language pairs. This will enable the analysis on the similarity of speech characteristics between different speakers and improve the transfer between languages.

%% file: references.tex
\bibliographystyle{template/lrec2022-bib}
\bibliography{references}

%% file: main.bbl
\begin{thebibliography}{}

\bibitem[\protect\citename{Andreeva \bgroup et al.\egroup }2014]{pitch2}
Andreeva, B., Demenko, G., Möbius, B., Zimmerer, F., Jügler, J., and
  Oleskowicz-Popiel, M.
\newblock (2014).
\newblock Differences of pitch profiles in germanic and slavic languages.

\bibitem[\protect\citename{Bahdanau \bgroup et al.\egroup
  }2016]{bahdanau2016neural}
Bahdanau, D., Cho, K., and Bengio, Y.
\newblock (2016).
\newblock Neural machine translation by jointly learning to align and
  translate.

\bibitem[\protect\citename{Beilharz \bgroup et al.\egroup }2020]{beilharz19}
Beilharz, B., Sun, X., Karimova, S., and Riezler, S.
\newblock (2020).
\newblock Librivoxdeen: A corpus for german-to-english speech translation and
  speech recognition.
\newblock {\em Proceedings of the Language Resources and Evaluation
  Conference}.

\bibitem[\protect\citename{Cattoni \bgroup et al.\egroup
  }2021]{CATTONI2021101155}
Cattoni, R., {Di Gangi}, M.~A., Bentivogli, L., Negri, M., and Turchi, M.
\newblock (2021).
\newblock Must-c: A multilingual corpus for end-to-end speech translation.
\newblock {\em Computer Speech \& Language}, 66:101155.

\bibitem[\protect\citename{Chen \bgroup et al.\egroup
  }2020]{chen2020multispeech}
Chen, M., Tan, X., Ren, Y., Xu, J., Sun, H., Zhao, S., Qin, T., and Liu, T.-Y.
\newblock (2020).
\newblock Multispeech: Multi-speaker text to speech with transformer.

\bibitem[\protect\citename{Do \bgroup et al.\egroup }2016]{porosody_estimate}
Do, T., Toda, T., Neubig, G., Sakti, S., and Nakamura, S.
\newblock (2016).
\newblock Preserving word-level emphasis in speech-to-speech translation.
\newblock {\em IEEE/ACM Transactions on Audio, Speech, and Language
  Processing}, PP.

\bibitem[\protect\citename{Dou and Neubig}2021]{awesomealign}
Dou, Z.-Y. and Neubig, G.
\newblock (2021).
\newblock Word alignment by fine-tuning embeddings on parallel corpora.
\newblock In {\em Conference of the European Chapter of the Association for
  Computational Linguistics (EACL)}.

\bibitem[\protect\citename{Eck \bgroup et al.\egroup }2010]{Jibbigo}
Eck, M., Lane, I., Zhang, Y., and Waibel, A.
\newblock (2010).
\newblock Jibbigo: Speech-to-speech translation on mobile devices.
\newblock In {\em 2010 IEEE Spoken Language Technology Workshop}, pages
  165--166.

\bibitem[\protect\citename{Giorgino}2009]{DTW}
Giorgino, T.
\newblock (2009).
\newblock Computing and visualizing dynamic time warping alignments in r: The
  dtw package.
\newblock {\em Journal of Statistical Software, Articles}, 31(7):1--24.

\bibitem[\protect\citename{Hutchins}2003]{hutchins2003development}
Hutchins, W.~J.
\newblock (2003).
\newblock {\em The development and use of machine translation systems and
  computer-based translation tools}.
\newblock Bahri.

\bibitem[\protect\citename{Ito and Johnson}2017]{ljspeech17}
Ito, K. and Johnson, L.
\newblock (2017).
\newblock The lj speech dataset.
\newblock \url{https://keithito.com/LJ-Speech-Dataset/}.

\bibitem[\protect\citename{Kocabiyikoglu \bgroup et al.\egroup
  }2018]{kocabiyikoglu2018augmenting}
Kocabiyikoglu, A.~C., Besacier, L., and Kraif, O.
\newblock (2018).
\newblock Augmenting librispeech with french translations: A multimodal corpus
  for direct speech translation evaluation.

\bibitem[\protect\citename{Kurimo \bgroup et al.\egroup
  }2010]{kurimo-etal-2010-personalising}
Kurimo, M., Byrne, W., Dines, J., Garner, P.~N., Gibson, M., Guan, Y.,
  Hirsim{\"a}ki, T., Karhila, R., King, S., Liang, H., Oura, K., Saheer, L.,
  Shannon, M., Shiota, S., and Tian, J.
\newblock (2010).
\newblock Personalising speech-to-speech translation in the {EMIME} project.
\newblock In {\em Proceedings of the {ACL} 2010 System Demonstrations}, pages
  48--53, Uppsala, Sweden. Association for Computational Linguistics.

\bibitem[\protect\citename{Latorre \bgroup et al.\egroup
  }2018]{multispeakertts}
Latorre, J., Lachowicz, J., Lorenzo{-}Trueba, J., Merritt, T., Drugman, T.,
  Ronanki, S., and Viacheslav, K.
\newblock (2018).
\newblock Effect of data reduction on sequence-to-sequence neural {TTS}.
\newblock {\em CoRR}, abs/1811.06315.

\bibitem[\protect\citename{Lewis}2015]{lewis2015skype}
Lewis, W.~D.
\newblock (2015).
\newblock Skype translator: Breaking down language and hearing barriers.
\newblock {\em Translating and the Computer (TC37)}, 10:125--149.

\bibitem[\protect\citename{Lo \bgroup et al.\egroup }2019]{mosnet}
Lo, C., Fu, S., Huang, W., Wang, X., Yamagishi, J., Tsao, Y., and Wang, H.
\newblock (2019).
\newblock Mosnet: Deep learning based objective assessment for voice
  conversion.
\newblock {\em CoRR}, abs/1904.08352.

\bibitem[\protect\citename{McAuliffe \bgroup et al.\egroup }2017]{mfa}
McAuliffe, M., Socolof, M., Mihuc, S., Wagner, M., and Sonderegger, M.
\newblock (2017).
\newblock Montreal forced aligner: Trainable text-speech alignment using kaldi.
\newblock In {\em INTERSPEECH}, pages 498--502.

\bibitem[\protect\citename{MircosoftAzure}2021]{microsoftazure}
MircosoftAzure.
\newblock (2021).
\newblock Speech translation.
\newblock
  \url{https://azure.microsoft.com/en-us/services/cognitive-services/speech-translation/#features}.

\bibitem[\protect\citename{Müller}2019]{thorsten}
Müller, T.
\newblock (2019).
\newblock Thorsten: Open german voice dataset.
\newblock
  \url{https://github.com/thorstenMueller/deep-learning-german-tts#dataset-thorsten-neutral}.

\bibitem[\protect\citename{Nakamura}2009]{nakamura2009overcoming}
Nakamura, S.
\newblock (2009).
\newblock Overcoming the language barrier with speech translation technology.
\newblock Technical report, Citeseer.

\bibitem[\protect\citename{Niebuhr and Skarnitzl}2019]{pitch}
Niebuhr, O. and Skarnitzl, R.
\newblock (2019).
\newblock Measuring a speaker's acoustic correlates of pitch -but which? a
  contrastive analysis based on perceived speaker charisma.

\bibitem[\protect\citename{Panayotov \bgroup et al.\egroup }2015]{librispeech}
Panayotov, V., Chen, G., Povey, D., and Khudanpur, S.
\newblock (2015).
\newblock Librispeech: An asr corpus based on public domain audio books.
\newblock In {\em 2015 IEEE International Conference on Acoustics, Speech and
  Signal Processing (ICASSP)}, pages 5206--5210.

\bibitem[\protect\citename{Pettarin}2017]{pettarin_2017}
Pettarin, A.
\newblock (2017).
\newblock aeneas.
\newblock \url{https://www.readbeyond.it/aeneas/}.

\bibitem[\protect\citename{Post \bgroup et al.\egroup }2013]{post2013improved}
Post, M., Kumar, G., Lopez, A., Karakos, D., Callison-Burch, C., and Khudanpur,
  S.
\newblock (2013).
\newblock Improved speech-to-text translation with the {F}isher and {C}allhome
  {S}panish--{E}nglish speech translation corpus.
\newblock In {\em Proceedings of the International Workshop on Spoken Language
  Translation (IWSLT)}, Heidelberg, Germany.

\bibitem[\protect\citename{Ren \bgroup et al.\egroup }2019]{ren2019fastspeech}
Ren, Y., Ruan, Y., Tan, X., Qin, T., Zhao, S., Zhao, Z., and Liu, T.-Y.
\newblock (2019).
\newblock Fastspeech: Fast, robust and controllable text to speech.

\bibitem[\protect\citename{Ren \bgroup et al.\egroup }2021]{ren2021fastspeech}
Ren, Y., Hu, C., Tan, X., Qin, T., Zhao, S., Zhao, Z., and Liu, T.-Y.
\newblock (2021).
\newblock Fastspeech 2: Fast and high-quality end-to-end text to speech.

\bibitem[\protect\citename{Sečujski \bgroup et al.\egroup
  }2016]{inproceedings}
Sečujski, M., Gerazov, B., Csapó, T., Delić, V., Garner, P., Gjoreski, A.,
  Guennec, D., Ivanovski, Z., Melov, A., Németh, G., Stojkovikj, A., and
  Szaszák, G.
\newblock (2016).
\newblock Design of a speech corpus for research on cross-lingual prosody
  transfer.
\newblock pages 199--206.

\bibitem[\protect\citename{Shen \bgroup et al.\egroup }2018]{shen2018natural}
Shen, J., Pang, R., Weiss, R.~J., Schuster, M., Jaitly, N., Yang, Z., Chen, Z.,
  Zhang, Y., Wang, Y., Skerry-Ryan, R., Saurous, R.~A., Agiomyrgiannakis, Y.,
  and Wu, Y.
\newblock (2018).
\newblock Natural tts synthesis by conditioning wavenet on mel spectrogram
  predictions.

\bibitem[\protect\citename{Skerry-Ryan \bgroup et al.\egroup
  }2018]{skerryryan2018endtoend}
Skerry-Ryan, R., Battenberg, E., Xiao, Y., Wang, Y., Stanton, D., Shor, J.,
  Weiss, R.~J., Clark, R., and Saurous, R.~A.
\newblock (2018).
\newblock Towards end-to-end prosody transfer for expressive speech synthesis
  with tacotron.

\bibitem[\protect\citename{Sperber and Paulik}2020]{sperber2020speech}
Sperber, M. and Paulik, M.
\newblock (2020).
\newblock Speech translation and the end-to-end promise: Taking stock of where
  we are.

\bibitem[\protect\citename{Wahlster}2013]{wahlster2013verbmobil}
Wahlster, W.
\newblock (2013).
\newblock {\em Verbmobil: Foundations of Speech-to-Speech Translation}.
\newblock Artificial Intelligence. Springer Berlin Heidelberg.

\bibitem[\protect\citename{Wang \bgroup et al.\egroup }2017]{wang2017tacotron}
Wang, Y., Skerry-Ryan, R., Stanton, D., Wu, Y., Weiss, R.~J., Jaitly, N., Yang,
  Z., Xiao, Y., Chen, Z., Bengio, S., Le, Q., Agiomyrgiannakis, Y., Clark, R.,
  and Saurous, R.~A.
\newblock (2017).
\newblock Tacotron: Towards end-to-end speech synthesis.

\bibitem[\protect\citename{Wang \bgroup et al.\egroup }2020]{wang2020covost}
Wang, C., Wu, A., and Pino, J.
\newblock (2020).
\newblock Covost 2 and massively multilingual speech-to-text translation.

\bibitem[\protect\citename{Wang \bgroup et al.\egroup }2021]{wang2021voxpopuli}
Wang, C., Rivi{\`e}re, M., Lee, A., Wu, A., Talnikar, C., Haziza, D.,
  Williamson, M., Pino, J., and Dupoux, E.
\newblock (2021).
\newblock Voxpopuli: A large-scale multilingual speech corpus for
  representation learning, semi-supervised learning and interpretation.
\newblock {\em arXiv preprint arXiv:2101.00390}.

\bibitem[\protect\citename{Yang \bgroup et al.\egroup }2020]{mbmel}
Yang, G., Yang, S., Liu, K., Fang, P., Chen, W., and Xie, L.
\newblock (2020).
\newblock Multi-band melgan: Faster waveform generation for high-quality
  text-to-speech.
\newblock {\em CoRR}, abs/2005.05106.

\bibitem[\protect\citename{Zanon~Boito \bgroup et al.\egroup
  }2020]{zanon-boito-etal-2020-mass}
Zanon~Boito, M., Havard, W., Garnerin, M., Le~Ferrand, {\'E}., and Besacier, L.
\newblock (2020).
\newblock {M}a{SS}: A large and clean multilingual corpus of sentence-aligned
  spoken utterances extracted from the {B}ible.
\newblock In {\em Proceedings of the 12th Language Resources and Evaluation
  Conference}, pages 6486--6493, Marseille, France. European Language Resources
  Association.

\end{thebibliography}
